%% file: main.tex
\pdfoutput=1

\documentclass[11pt]{article}

\usepackage{naacl2021}

\usepackage{times}
\usepackage{latexsym}

\usepackage[T1]{fontenc}

\usepackage[utf8]{inputenc}

\usepackage{microtype}

\usepackage{xspace}
\usepackage{xcolor}
\usepackage{graphicx}
\usepackage[normalem]{ulem}
\usepackage{amsmath}
\usepackage{enumitem}

\usepackage{hyperref}
\usepackage{breakurl}

\usepackage{booktabs}
\usepackage{multirow}
\usepackage{makecell}
\usepackage{ragged2e}

\usepackage{caption}
\usepackage{subcaption}

\usepackage{listings}

\usepackage{tikz}
\usetikzlibrary{arrows}
\usetikzlibrary{calc}
\usetikzlibrary{shapes.geometric}
\usetikzlibrary{decorations.pathreplacing}
\usetikzlibrary{positioning}

\usepackage{svg}
\usepackage[switch]{lineno}

\newcommand{\XSpace}[1]{}
\newcommand{\XComment}[1]{}

\newcommand{\DefMacro}[2]{\expandafter\newcommand\csname rmk-#1\endcsname{#2}}
\newcommand{\UseMacro}[1]{\csname rmk-#1\endcsname}

\newcommand\redsout{\bgroup\markoverwith{\textcolor{red}{\rule[0.5ex]{2pt}{0.4pt}}}\ULon}

\newcommand{\MyPara}[1]{\vspace{2pt}\noindent\textbf{#1}.}

\newcommand{\InputWithSpace}[1]{\bgroup\def\arraystretch{1.0}\input{#1}\egroup}
\newcommand{\Code}[1]{{\ifmmode{\mathtt{#1}}\else$\mathtt{#1}$\fi}}
\newcommand{\CodeIn}[1]{{\ifmmode{\mathtt{#1}}\else$\mathtt{#1}$\fi}}

\newcolumntype{R}[1]{>{\RaggedLeft\arraybackslash}p{#1}}
\newcolumntype{L}[1]{>{\RaggedRight\arraybackslash}p{#1}}

\definecolor{gray}{RGB}{211,211,211}
\definecolor{javared}{rgb}{0.6,0,0} %
\definecolor{javagreen}{rgb}{0.25,0.5,0.35} %
\definecolor{javapurple}{rgb}{0.5,0,0.2} %
\definecolor{javadocblue}{rgb}{0.25,0.35,0.75} %
\newcommand{\jbasicstyle}{\small\sffamily} %

\newcommand{\jnumberstyle}{\scriptsize}

\lstdefinelanguage{pseudo}
{
  morekeywords={},
  keywordstyle=\bfseries,
  lineskip=-0.1em,
  numbers=left, %
  numberstyle=\jnumberstyle,
  numbersep=4pt,
  basicstyle=\jbasicstyle,
  breaklines=true,
  breakautoindent=true,
  tabsize=2,
  columns=fullflexible,
  morecomment=*[l][\textsl]{//},
  mathescape=true,
  xleftmargin=10pt,
}

\lstdefinelanguage{todo-comment}
{
  morekeywords={},
  keywordstyle=\bfseries,
  lineskip=-0.1em,
  numbers=none,
  basicstyle=\jbasicstyle,
  breaklines=true,
  breakautoindent=true,
  tabsize=2,
  columns=fullflexible,
  morecomment=*[l][\textsl]{//},
  mathescape=true,
  xleftmargin=-10pt,
}

\lstdefinelanguage{java-pretty}
{
  language=java,
  numbers=left,
  basicstyle=\scriptsize\ttfamily,
  numberstyle=\scriptsize,
  breaklines=true,
  columns=fullflexible,
  xleftmargin=16pt,
  showstringspaces=false,
  keywordstyle=\color{javapurple}\bfseries,
  stringstyle=\color{javared},
  commentstyle=\color{javagreen},
  morecomment=[s][\color{javadocblue}]{/**}{*/},
}

\newcommand{\Title}{Learning to Generate Code Comments from Class Hierarchies}

\newcommand{\subclass}{subclass\xspace}

\newcommand{\superclass}{superclass\xspace}
\newcommand{\Superclass}{Superclass\xspace}

\newcommand{\overrides}{overrides\xspace}
\newcommand{\overriding}{overriding\xspace}
\newcommand{\overridden}{overridden\xspace}
\newcommand{\niwf}{NIWF\xspace}
\newcommand{\addonesmoothing}{add-one smoothing\xspace}
\newcommand{\compatibility}{conformity\xspace}
\newcommand{\compatible}{conforms\xspace}
\newcommand{\incompatible}{non-conforming\xspace}

\newcommand{\ChildClassName}{\ensuremath{Kname^{\bot}}}
\newcommand{\ChildComment}{\ensuremath{C^{\bot}}}
\newcommand{\ParentComment}{\ensuremath{C^{\top}}}
\newcommand{\ParentMethod}{\ensuremath{M^{\top}}}
\newcommand{\ChildMethod}{\ensuremath{M^{\bot}}}
\newcommand{\ChildClass}{\ensuremath{K^{\bot}}}
\newcommand{\ParentClass}{\ensuremath{K^{\top}}}
\newcommand{\ParentClassName}{\ensuremath{Kname^{\top}}}
\newcommand{\SeqToSeq}{\textsc{Seq2Seq}\xspace}
\newcommand{\Copy}{\textsc{Copy}\xspace}
\newcommand{\DeepCom}{\textsc{DeepCom-hybrid}\xspace}
\newcommand{\ClassNameSubst}{\textsc{Class Name Substitution}\xspace}
\newcommand{\EditModel}{\textsc{Edit Model}\xspace}
\newcommand{\EditModelPlus}{\textsc{Edit Model +e-feats +rerank}\xspace}

\newcommand{\SuperAgnosticGen}{\SeqToSeq{}\xspace}
\newcommand{\SuperAwareGen}{\textsc{Hierarchy-Aware} \SeqToSeq{}\xspace}

\newcommand{\SuperAwareGenWithSpecRerank}{\SuperAwareGen{}\xspace}
\newcommand{\SuperAwareGenWOutUll}{-unlikelihood\xspace}
\newcommand{\SuperAwareGenWOutSpecWOutRerank}{-unlikelihood, specificity\xspace}
\newcommand{\SuperAwareGenWOutFeatWOutRerankWOutSpec}{--unlikelihood, specificity, feats\xspace}
\newcommand{\SuperAwareGenWOutClassName}{-unlikelihood, specificity, class
name encoder\xspace}
\newcommand{\SuperAwareGenWOutComment}{-unlikelihood, specificity, \superclass
comment encoder\xspace}

\newcommand{\SuperAwareGenWithAST}{\SuperAwareGen+AST\xspace}

\newcommand{\xshort}{first sentence\xspace}
\newcommand{\Short}{First sentence\xspace}
\newcommand{\xlong}{full description\xspace}
\newcommand{\Long}{Full description\xspace}

\newcommand{\SuperClassExample}{\CodeIn{Object}\xspace}
\newcommand{\SubClassExample}{\CodeIn{InfoAccessSyntax}\xspace}
\newcommand{\OverridenMethodExample}{\CodeIn{getEncoded()}\xspace}

\DefMacro{DeepCom}{\DeepCom{}}
\DefMacro{Copy}{\Copy{}}
\DefMacro{ClsSubt}{\ClassNameSubst{}}
\DefMacro{GenModel}{\SuperAgnosticGen{}}
\DefMacro{EditModel}{\EditModel{}}
\DefMacro{CopyGen}{}
\DefMacro{CopyGen-spec}{CopyGen+spec}
\DefMacro{CopyGen-neg-remove-overlap}{\SuperAwareGen{}}

\newcommand{\Tstrut}{\rule{0pt}{2.6ex}}

\newcommand{\TableCaptionDatasetMetrics}{
  Number of projects, number of examples before preprocessing, and
  number of examples after preprocessing (in \xshort and \xlong
  datasets) in training, validation, and testing
  sets.\label{tab:dataset-metrics}}
\newcommand{\TableCaptionDatasetMetricsAux}{
  Average number of subtokens in \ParentComment{} and \ChildComment{}
  in \xshort and \xlong datasets.\label{tab:dataset-metrics-aux}}
\newcommand{\TableCaptionQualitativeExample}{
  Qualitative analysis of our \SuperAwareGen model and the baselines
  models' outputs on one example.\label{tab:qualitative-example}}
\newcommand{\TableCaptionQualitativeExampleApp}{
  Qualitative analysis of our \SuperAwareGen model and the baselines
  models' outputs on one example.\label{tab:qualitative-example-app}}
\newcommand{\TableCaptionModelsResults}{
  Comparison of our \SuperAwareGen model with the baseline methods.
  The differences between each pair of models on each metric are
  statistically significant.\label{tab:results}}
\newcommand{\TableCaptionAblation}{
  Ablation study of our \SuperAwareGen{} model. The differences
  between the models with the same Greek letter suffix (and only those
  pairs) are \textbf{not} statistically significant.\label{tab:ablt-study}}

\newcommand{\TableHeadNA}{\makecell[c]{n/a}\xspace}
\newcommand{\TableHeadNumProjects}{\#Projects\xspace}
\newcommand{\TableHeadBeforePreprocessing}{Before preprocessing\xspace}
\newcommand{\TableHeadAvgLenParentComment}{Avg. len(\ParentComment{})\xspace}
\newcommand{\TableHeadAvgLenChildComment}{Avg. len(\ChildComment{})\xspace}
\DefMacro{TableHead-train}{Train\xspace}
\DefMacro{TableHead-valid}{Valid\xspace}
\DefMacro{TableHead-test}{Test\xspace}
\DefMacro{TableHead-short}{\Short{}\xspace}
\DefMacro{TableHead-long}{\Long{}\xspace}

\input{tables/numbers-dataset-metrics.tex}
\input{tables/numbers-comlen.tex}
\input{tables/numbers-copy-results.tex}
\input{tables/numbers-deepcom-results.tex}
\input{tables/numbers-clssubt-results.tex}
\input{tables/numbers-genmodel-results.tex}
\input{tables/numbers-copygen-wo-ull-results.tex}

\input{tables/numbers-editmodel-results.tex}

\input{tables/numbers-copygen-results.tex}
\input{tables/numbers-ensemble-results.tex}
\input{tables/numbers-ablation-results.tex}

\input{tables/numbers-copygen-spec-results.tex}

\input{tables/numbers-copygen-neg-remove-overlap-results.tex}

\input{tables/numbers-copygencompatclfrerank-results.tex}

\input{tables/numbers-specificity.tex}

\input{tables/numbers-compat-clf.tex}

\input{tables/numbers-ast-copygencompatclfrerank-results.tex}

\input{tables/numbers-editmodel-feat-rerank-results.tex}

\newcommand{\NumTrial}{three\xspace}

\graphicspath{ {figures/} }

\title{\Title}

\author{
  Jiyang Zhang, Sheena Panthaplackel, Pengyu Nie, \\
  {\bf Raymond J. Mooney, Junyi Jessy Li, Milos Gligoric} \\
  The University of Texas at Austin, USA \\
  \texttt{\{jiyang.zhang@, spantha@cs., pynie@,}\\ \texttt{mooney@cs., jessy@austin., gligoric@\}utexas.edu} \\
}

\begin{document}
\maketitle

\begin{abstract}
  Descriptive code comments are essential for supporting code
  comprehension and maintenance. We propose the task of automatically generating comments for \textit{overriding} methods. We formulate a novel framework which accommodates the unique contextual and linguistic reasoning that is required for performing this task. Our approach features: (1)~incorporating context from the class hierarchy; (2)~conditioning on learned, latent representations of \textit{specificity} to generate comments that capture the more specialized behavior of the overriding method; and (3)~\textit{unlikelihood training} to discourage predictions which do not conform to invariant characteristics of the comment corresponding to the overridden method.
  Our experiments show that the proposed approach is able to generate
  comments for \overriding methods of higher quality compared to
  prevailing
comment generation techniques.
\end{abstract}

\section{Introduction}
\label{sec:intro}

Developers rely on natural language comments to understand key
aspects of the source code they accompany, such as implementation,
functionality, and usage. Comments are critical in supporting code
comprehension and maintenance. However, writing them can be
laborious and time-consuming, which is not ideal for fast-development
cycles. Unsurprisingly, this
has sparked interest in \textit{automatic comment
  generation}~\cite{iyer-etal-2016-summarizing,AhmadETAL20Transformer-based,
  clement-etal-2020-pymt5,YuEtal20Towards}.

Prior work on automatic comment generation study various comment types
including class-level comments~\cite{HaiducETAL10Automated,
  MorenoETAL13Automatic} and method-level
comments~\cite{SridharaETAL10GeneratingSummaryComments,
  iyer-etal-2016-summarizing, LiangAndZhu18Automatic, HuETAL19Deep,
  LeClairETAL20Improved, FernandesETAL19Structured, LiuETAL20Automating}.
A large portion of method-level comments correspond to \overriding methods.
Figure~\ref{fig:example} shows an example of method \overriding, where
the \SubClassExample class \emph{inherits} characteristics (methods
and fields) from its \superclass{} \SuperClassExample, and
\SubClassExample.\OverridenMethodExample \overrides{}
\SuperClassExample.\OverridenMethodExample.
\citet{TemperoETAL10Emperical} find that half of the projects they
inspected have 53\% or more of their subclasses overriding at least
one method from the superclass.

Automatic comment generation for \overriding methods poses unique challenges.
In the example in
Figure~\ref{fig:example}, \SubClassExample.\OverridenMethodExample
adds the functionality to initialize the encoding, i.e., when the
encoding is \CodeIn{null}, the method will compute and use the
\CodeIn{ASN1} encoding.  Generating a comment for the
overriding method, \SubClassExample.\OverridenMethodExample,
requires reasoning about how the method relates to the
overridden method, \SuperClassExample.\OverridenMethodExample,
and addressing the customization that is done.
In this paper, we formulate the task of generating comments for
\overriding methods and propose an approach, \SuperAwareGen, for
facilitating the complex contextual and linguistic reasoning that is
needed for performing this task.

\SuperAwareGen leverages contextual information from multiple components of the class hierarchy.
Namely, we incorporate a learned representation of the
comment corresponding to the \overridden method in the \superclass, as
this provides a general template that can be adapted. We also encode
the class name, which often sheds light on the customization and
sometimes
appears in the comment, as seen in Figure~\ref{fig:example}
(\SubClassExample in the comment).

Relative to comments for \overridden methods, we observe that comments
accompanying \overriding methods use more specific language to
describe the specialized functionality of the overriding method (e.g.,
\emph{``ASN.1''} in Figure~\ref{fig:example}). However,
encoder-decoder models naturally prefer tokens with higher frequency
in the data during decoding, a known phenomena in dialog
generation~\cite{sordoni-etal-2015-neural, SerbanETAL16Building,
  li-etal-2016-diversity} that causes a preference for generic
language.  To encourage predictions which are more specific in nature,
we factor out the notion of
\emph{specificity}~\cite{zhang-etal-2018-learning-control,ko-etal-2019-linguistically}
in our model and condition on learned representations of specificity
when generating comments.
Additionally, we capture the intuition that the \overriding comment
should retain key aspects of the \overridden comment that are not
affected by the customization done in the \overriding method. To do
this, we use \textit{unlikelihood training}~\cite{LiETAL20Dont,WelleckETAL19Neural} with
synthetically generated sequences that do not conform to invariant
characteristics of the \overridden comment.

\begin{figure}[t]
  \centering
  \begin{subfigure}{0.5\textwidth}
\begin{lstlisting}[language=java-pretty, numbers=none]
public class Object {
    protected byte[] encoding;
    /** Returns encoded form of the object. */
    public byte[] getEncoded() {
        return encoding;}}
\end{lstlisting}
    \caption{\OverridenMethodExample in the \superclass, \SuperClassExample.}
    \label{fig:parent}
  \end{subfigure}
  \begin{subfigure}{0.5\textwidth}
\begin{lstlisting}[language=java-pretty, numbers=none]
public class InfoAccessSyntax extends Object {
    /** Returns ASN.1 encoded form of this infoAccessSyntax. */
    @Override public byte[] getEncoded() {   
        if (encoding == null) {
            encoding = ASN1.encode(this);}
        return encoding;}}
\end{lstlisting}
    \caption{\OverridenMethodExample in the \subclass, \SubClassExample.}
    \label{fig:child}
  \end{subfigure}
  \caption{Class \SubClassExample{} extends the
    \superclass{} \SuperClassExample{} and \overrides{} the
    \OverridenMethodExample{} method.\label{fig:example}}
\end{figure}

For training and evaluation, we construct a parallel corpus
consisting of \UseMacro{long-train-numbers} Java
\overriding-\overridden method-comment tuples, with the corresponding
details regarding class hierarchy.
Experimental results show that our approach outperforms multiple baselines, including prevailing
models for comment generation~\cite{HuETAL19Deep} and comment update~\cite{PanthaplackelETAL20Learning}
by significant margins.
Through an ablation study, we
demonstrate the effectiveness
of the designed class hierarchy context, specificity-factored
architecture, and unlikelihood training for \compatibility, where all of them
statistically significantly improved the generation quality.

We summarize our main contributions as follows: (1)~we formulate the
novel task of generating comments for \overriding methods;  (2)~to
address this task, we design an approach
that leverages aspects of the class hierarchy to gather broader
contextual information as well as to learn notions of specificity and
\compatibility, and we show that this approach can outperform multiple
baselines;
(3)~for training and evaluation, we build a large corpus of
\overriding-\overridden method-comment tuples, with their associated
class hierarchy information.
We will make our code and data publicly
available upon publication.

\section{Task}
\label{sec:task}

When a developer overrides a method, we aim to automatically generate a
natural language comment which accurately reflects the method's behavior,
capturing important aspects of the class hierarchy as well as the
customization that extends the implementation in the \superclass.
Concretely, in Figure~\ref{fig:example}, suppose a developer writes the
method body of \ChildMethod{} (\OverridenMethodExample) in class
\ChildClass{} (\SubClassExample) which \overrides the parent method,
\ParentMethod{}, in the \superclass, \ParentClass{}
(\SuperClassExample). Our task is to generate the comment for
\ChildMethod{}, \ChildComment{} (\textit{``Returns ASN.1 encoded form
  of this infoAccessSyntax''}), using context provided by the
\overriding method body, \ChildMethod{}, as well as the inputs from
the class hierarchy.
This includes the name of \ChildClass{} (\ChildClassName{}), the name of \ParentClass{} (\ParentClassName{}), the
\overridden method body (\ParentMethod{}), and \ParentComment{} (i.e.,
\ParentMethod{}'s comment, \textit{``Returns encoded form of the
object''}).

\section{\SuperAwareGen}
\label{sec:CopyGen}
Our approach, \SuperAwareGen, leverages context from the class
hierarchy to generate \ChildComment{}.
Depicted in Figure~\ref{fig:models}, this is a
\SeqToSeq~\cite{Sutskever2014Seq2Seq} model that decodes
\ChildComment{} using learned representations from three encoders:
\ChildMethod{} encoder (Section~\ref{sec:method_encoder}),
\ChildClassName{} encoder (Section~\ref{sec:class-name-encoder}), and
\ParentComment{} encoder
(Section~\ref{sec:parent-comment-encoder}). We also incorporate
token-level auxiliary features into each of these encoders in order to
further capture patterns pertaining to the class hierarchy, as well as
properties of code and comments. Next, we discourage generating generic predictions by additionally injecting tailored
representations that capture specificity (Section~\ref{sec:specificity}). Finally, we use unlikelihood training to discourage the model from making predictions which fail to capture key aspects of \ParentComment{} that should persist in the generated comment for \ChildMethod{}
(Section~\ref{sec:compat-reranking}).

\begin{figure*}[t]
  \centering
  \resizebox{.75\linewidth}{!}{\input{figures/model-architecture}}
  \caption{The neural architecture of \SuperAwareGen.}
  \label{fig:models}
\end{figure*}

\subsection{Method Encoder}
\label{sec:method_encoder}
We use a bidirectional GRU~\cite{ChoETAL14Learning} encoder to learn a
representation for \ChildMethod{}. We concatenate auxiliary features
to the embedding vector corresponding to each input token before
feeding it into the encoder.  We include indicator features for
whether a token is a Java keyword or operator, to capture common
patterns among these types of tokens.
Next, we compute the \textit{diff} between \ParentMethod{} and
\ChildMethod{} in order to construct features identifying whether a
code token is retained, added, deleted, or replaced. By establishing
the similarities and differences, we aim to capture how the \subclass
implementation (\ChildMethod{}) relates to that of the \superclass
(\ParentMethod{}). In addition, we add the features indicating whether
the code tokens overlap with class names (\ChildClassName{} and
\ParentClassName{}) or \ParentComment{} as they usually contain
important tokens such as object names and field names which may appear
in the comments. We believe this will help guide the model in
determining how to utilize the context provided by other components of
the class hierarchy.

\subsection{Class Name Encoder}
\label{sec:class-name-encoder}
Because \ChildMethod{} is designed specifically for the \subclass
(\ChildClass{}), the class name (\ChildClassName{}) can often provide
deeper insight into \ChildMethod{}'s functionality.
For instance, in Figure~\ref{fig:example}, it is impossible to
generate the correct comment without context from \ChildClassName{}
(\SubClassExample).
Using a bidirectional GRU encoder, we learn a representation of
\ChildClassName{} as a sequence of subtokens (we split class names in
the form of CamelCase into subtokens).
Similarly, we extract the features identifying whether a token in
\ChildClassName{} is retained, added, deleted, or replaced compared
with \ParentClassName{}. Features which indicate the overlap between
\ChildClassName{} and \ChildMethod{} are included as well.

\subsection{\Superclass Comment Encoder}
\label{sec:parent-comment-encoder}
We find the hierarchical relationship between \ChildMethod{} and
\ParentMethod{} to often hold between their accompanying comments. 
Namely, similar to how \ChildMethod{} inherits the general
structure and functionality from \ParentMethod{}, \ChildComment{}
often follows the general format and shares some amount of content
with \ParentComment{}, as seen in Figure~\ref{fig:example}.
To provide context from the comment in the \superclass, we introduce
an encoder for learning a representation of \ParentComment{}. Like the
other encoders, we concatenate features to the embedding vector
corresponding to each token in the input sequence. We include features
capturing lexical overlap with \ChildClassName{} and \ChildMethod{}
as we expect similar patterns to emerge in \ChildComment{}.
We also incorporate features that have been used to characterize
comments in prior work~\citep{PanthaplackelETAL20Learning}: whether
the token appears more than once, is a stop word, and its
part-of-speech tag.

\subsection{Decoder}
We concatenate the final hidden states of each of the encoders to
initialize the GRU decoder. To further inform the decoder of the
context from the three inputs, we leverage attention~\cite{LuongETAL15Effective} over
the hidden states of all of these encoders. We additionally allow the
decoder to copy tokens related to the implementation and class
hierarchy from context provided by the inputs through a pointer
network~\cite{VinyalsPointer} over the hidden states of all three encoders.

\subsection{Conditioning on Specificity}
\label{sec:specificity}

To test the hypothesis that \ChildComment{} contains information
specific to \ChildMethod{}, we compare the specificity of
\ChildComment{} and \ParentComment{} using normalized inverse word
frequency (\niwf)~\cite{zhang-etal-2018-learning-control}. \niwf is a
frequency-based metric that computes the maximum of the Inverse Word
Frequency (IWF) of all the tokens $w$ in a comment $C$, normalized to 0--1:

\vspace{-10pt}
{\small
\begin{flalign*}
  \mathtt{max}(\mathtt{IWF}_C) = \mathtt{max}_{w\in C} \left( \frac{\mathtt{log}(1 + |Y|)}{1 + f_w} \right)
\end{flalign*}}%
where $|Y|$ is the number of comments in the corpus, $f_w$ denotes the
number of comments that contain the token $w$.  We apply
\addonesmoothing such that the metric can be computed when the comment
contains out-of-vocabulary tokens.

We found significant differences in terms of the distribution of \niwf
among \ChildComment{} and \ParentComment{}: the \niwf metric of
\xshort comments (the summary sentences of comments) is \UseMacro{sub-niwf-short-train-AVG} on average for
\ChildComment{} and is \UseMacro{sup-niwf-short-train-AVG} on average
for \ParentComment{}; the \niwf metric of \xlong comments (the entire main descriptions, including the summary sentences as well as low-level descriptions of the methods' functionality) is
\UseMacro{sub-niwf-long-train-AVG} on average for \ChildComment{} and
is \UseMacro{sup-niwf-long-train-AVG} on average for \ParentComment{}.
In both \xshort and \xlong cases, the \niwf of \ChildComment{} is
statistically significantly higher than that of \ParentComment{} using
paired Wilcoxon signed-rank tests~\cite{Wilcoxon45Individual} under
confidence level of 95\%.

To encourage the decoder to predict a sequence that is specific and concretely reflects the functionality of \ChildMethod{}, we follow prior work in dialog generation~\cite{ko-etal-2019-linguistically} 
and learn embeddings of specificity jointly with the model.
Namely, we discretize NIWF
into $K = 5$ levels of equal size, and associate an embedding with each level.
During training, each comment is assigned into its corresponding level, 
thus the embeddings are trained jointly with the model.
At test time, 
following the intuition that \overriding comments should be more
specific, we use the level that maximizes specificity,
and concatenate its embedding with the decoder input word embedding at each time step.\footnote{In our
  preliminary experiments, we also explored using
  $\mathtt{max}(\mathtt{specificity\_level}(\ParentComment{}) + 1, K)$
  at inference time (thus we used \addonesmoothing during computing
  \niwf such that it can be computed on \ParentComment{} in test set
  that may contain out-of-vocabulary words), but it resulted in worse
  performance.}

Because specificity alone would encourage the model to generate words
of lower frequency, we additionally encourage the model to prefer
tokens that are semantically similar to the input. Specifically, we
calculate coherence, which measures the similarity between the
\ChildComment{} and inputs representations; the representation of a
sentence ($\mathtt{repr}(\mathtt{s})$) is computed as the weighted
average of all word embeddings in $\mathtt{s}$, where the weight of
each token is its inverse document frequency (with \addonesmoothing):

\vspace{-10pt}
{\small
\begin{flalign*}
  \mathtt{repr}(\mathtt{s}) &= \sum_{w\in s} \left( \frac{e_w}{1 + f_w} \right) \\
  \mathtt{coherence}(\ChildComment{}, \mathtt{s}) &= \mathtt{cossim}(\mathtt{repr}(\ChildComment{}), \mathtt{repr}(\mathtt{s}))
\end{flalign*}}%
where $\mathtt{s}\in\{ \ParentComment{}, \ChildMethod{},
\ChildClassName{} \}$, $\mathtt{cossim}$ computes the cosine
similarity, and $e_w$ is the word embedding of $w$.
Similar to specificity, these coherence representations are also discretized into 5 levels, and their embeddings jointly trained with the model. At inference time, we select the maximum level for coherence.

\subsection{Unlikelihood Training}
\label{sec:compat-reranking}

We additionally ensure that the model prediction \ChildComment{}
\compatible to invariant characteristics of \ParentComment{}. In other
words, the model prediction should be suitable for the \overriding
method while simultaneously retaining the pertinent content and
structure of the comment accompanying the \overridden method. To
discourage the model from generating sequences that do not preserve
the salient information, we incorporate an unlikelihood objective into
our training procedure~\cite{LiETAL20Dont,WelleckETAL19Neural}. Essentially, the model is trained to assign
low probability scores to \incompatible sequences which do not contain
all tokens that should be retained from \ParentComment{}. For each
example in the training set, we synthetically generate a \incompatible
instance by randomly removing 10\% of \ChildComment{}'s tokens which
also appear in \ParentComment{} (if the fraction of such tokens is
less than 10\%, after removing them, we randomly remove remaining
tokens until 10\% of \ChildComment{}'s tokens are removed). We refer
to the examples in our standard training set as \textit{positive}
examples and the synthetically generated \incompatible cases as
\textit{negative} examples.

During training, the standard likelihood loss can be calculated for each positive example:
\begin{flalign*}
  \scriptstyle L_{MLE}^{(i)}(p_{\theta}, x^{(i)}, y^{(i)+}) = -\sum_{t=1}^{|y^{(i)+}|}\log{p_{\theta}(y_{t}^{(i)+}| x^{(i)}, y^{(i)+}_{1:t-1})},
\end{flalign*}
where $x^{(i)}$ is the input contexts, $y^{(i)+}$ is the positive target,
and $y_{t}^{(i)+}$ is the $t$-th token of $y^{(i)+}$.  Meanwhile, the unlikelihood loss
can be calculated on the corresponding negative example:
\begin{flalign*}
\scriptstyle L_{UL}^{(i)}(p_{\theta}, x^{(i)}, y^{(i)-}) = -\sum_{t=1}^{|y^{(i)-}|}\log{(1 - p_{\theta}(y_{t}^{(i)-}| x^{(i)}, y^{(i)-}_{1:t-1}))},
\end{flalign*}
where $y^{(i)-}$ is the generated negative target.
The overall loss function consists of mixing of likelihood and
unlikelihood losses:
\begin{flalign*}
\scriptstyle L_{ULE}^{(i)} = L_{MLE}^{(i)} + \alpha L_{UL}^{(i)} 
\end{flalign*}
where $\alpha$ is the mixing hyper-parameter.

\section{Dataset}
\label{sec:dataset}

\input{tables/table-dataset-metrics.tex}

\input{tables/table-dataset-metrics-aux.tex}

Because prior work do not consider class hierarchy, existing
datasets do not include necessary information for our models and
task.
Therefore, we build a new corpus by mining open-source Java
projects for (method, comment) pairs along with information from their
class hierarchy. From the same projects that comprise a commonly used
comment generation dataset~\cite{HuETAL18Deep}, we extract examples in
the form: ((\ChildMethod{}, \ChildComment{}, \ChildClass{}),
(\ParentMethod{}, \ParentComment{}, \ParentClass{})). From the Javadoc
API documentation accompanying a given method, we derive comments from
the main description, which precedes the
tags \footnote{\url{https://www.oracle.com/technical-resources/articles/java/javadoc-tool.html\#tag}},
e.g., \CodeIn{@param}.

We specifically consider generating the \emph{\xshort} of the
main description, similar to prior work~\cite{HuETAL19Deep}. The
first sentence serves as a summary comment of the high-level
functionality of the method. We also consider the more
challenging task: generating the \emph{\xlong} comment, i.e., the
entire main description.
The \xlong comment includes both the high-level summary, as well as low-level
description of the method's functionality.

\MyPara{Preprocessing}
Similar to what was done in prior
work~\cite{Movshovitz-AttiasCohen13PredictingProgrammingComments,
  PanthaplackelETAL20Learning}, we partition the dataset in such a way
that there is no overlap between the projects used for training,
validation, and testing.  By limiting the number of closely-related
examples (i.e, methods from sibling classes) across partitions, this
setting allows us to better evaluate a model's ability to generalize.
We filter out comments with non-English words, and remove those with
$<3$ words, as we find that these often fail to adequately describe
functionality.  We also discard trivial examples in which
\ChildComment{}=\ParentComment{} as they lead to unwanted behavior in
which the model learns to just copy \ParentComment{}.  Finally, we
tokenize source code and comments by splitting by space and
punctuation and then split tokens of the form \textit{camelCase} and
\textit{snake\_case} to subtokens in order to reduce vocabulary
size. 

\MyPara{Statistics} We show statistics on our corpus before and after
preprocessing in Table~\ref{tab:dataset-metrics}.
The number of examples for the first sentence is smaller than that of
the full description because there are more instances of
\ChildComment{}=\ParentComment{} for first sentence, which needed to
be filtered. This is expected, as they are shorter and more high-level
in nature, allowing them to be the same when \ChildMethod{} and
\ParentMethod{} have the same high-level description.
Furthermore, while there are fewer examples in the test sets, in
comparison to the validation sets, the number of projects from which
the test examples come from is significantly larger than that of the
validation examples. This is an artifact of using the cross-project
setting, resulting from differences in the distribution of examples
across various projects.  Table~\ref{tab:dataset-metrics-aux} shows
the average number of subtokens in \ParentComment{} and
\ChildComment{} in our datasets.

\section{Experiments}
\label{sec:eval}

In this section, we describe several baselines, implementation
details, evaluation metrics, and experiment setup.

\subsection{Baselines}
\label{sec:baselines}

We compare our model with two rule-based baselines (\Copy{} and \ClassNameSubst{}), a generation model without using class hierarchy (\SuperAgnosticGen), and two prevailing deep learning methods (\DeepCom{}~\cite{HuETAL19Deep} and \EditModel{}~\cite{PanthaplackelETAL20Learning}).

\MyPara{\Copy} This is a rule-based approach which merely
copies \ParentComment as the prediction for \ChildComment{}. This resembles the functionality of the Javadoc tool.
\footnote{\url{https://www.oracle.com/technical-resources/articles/java/javadoc-tool.html\#reusingcomments}}

\MyPara{\ClassNameSubst} Based on our observations, there are many cases in which the developer
obtains \ChildComment{} by simply copying \ParentComment{} and
replacing all occurrences of the parent class name with that of the
child class. We simulate this procedure using rule-based string
replacement: \ChildComment{} =
\ParentComment{}.\texttt{Replace}(\ParentClass{}, \ChildClass{}). Note that
if the parent class name does not appear in \ParentComment{}, then
\ChildComment{}=\ParentComment{}.

\MyPara{\SuperAgnosticGen}
In line with prior work that generates \ChildComment{} using
\ChildMethod{} alone~\cite{iyer-etal-2016-summarizing}, we consider an
approach that does not have access to class-related information. This
approach is a baseline version of \SuperAwareGen
model without the class name, \superclass comment encoders, without conditioning on specificity, and
unlikelihood training. We also disregard
the auxiliary features used in the method encoder as these include
attributes of the class hierarchy.

\MyPara{\DeepCom}~\cite{HuETAL19Deep} This approach
generates a comment for a given Java method by learning the
representations of the corresponding abstract syntax tree (AST) and
code sequence respectively. They flatten the AST into a sequence using
the Structure-based Traversal (SBT) algorithm. The two sequences are encoded with separate long short-term memory (LSTM) networks.
An LSTM decoder
then uses these learned representations in order to generate a
sequence of comment tokens.

\MyPara{\EditModel} We find that developers often produce \ChildComment{} by editing
\ParentComment{}; however, these are not always as simple as class
name substitution and require more complex edits. To address this, we
include a model which \textit{learns} to edit \ParentComment{} in
order to produce \ChildComment{}.
We adapt a recent comment editing framework that was originally
proposed for updating comments that become outdated upon code changes
to the corresponding methods~\cite{PanthaplackelETAL20Learning}. They
first encode the existing comment using a bidirectional GRU encoder
and the code edits with another bidirectional GRU encoder. They then
use a GRU decoder to generate a sequence of comment edits which are
applied to the existing comment. This leads to an updated comment that
is consistent with the new version of the method. In our setting, we
treat \ParentComment{} as the ``existing comment" and encode the code
edits between \ParentMethod{} and \ChildMethod{}. We apply the
generated comment edit sequence to \ParentComment{} in order to
produce \ChildComment{}, which is expected to be consistent with
\ChildMethod{}.
Additionally, we use the same auxiliary features and reranking
mechanism they originally proposed.

\begin{table*}
\small
\centering
\begin{tabular}{c|rrr|rrr}
\toprule
\multirow{2}{*}{\textbf{Model}}
  &
  \multicolumn{3}{c|}{\Short{}}
  &
  \multicolumn{3}{c}{\Long{}}
  \\
  \cline{2-7}
 & \textbf{BLEU-4} \Tstrut & \textbf{METEOR} &\textbf{ROUGE-L}
& \textbf{BLEU-4} & \textbf{METEOR} &\textbf{ROUGE-L}\\
\midrule
\UseMacro{Copy}
 & \UseMacro{Copy-short-bleu}
 & \UseMacro{Copy-short-meteor}
 & \UseMacro{Copy-short-rouge-l}
 & \UseMacro{Copy-long-bleu}
 & \UseMacro{Copy-long-meteor}
 & \UseMacro{Copy-long-rouge-l}
\\
\UseMacro{ClsSubt}
 & \UseMacro{ClsSubt-short-bleu}
 & \UseMacro{ClsSubt-short-meteor}
 & \UseMacro{ClsSubt-short-rouge-l}
 & \UseMacro{ClsSubt-long-bleu}
 & \UseMacro{ClsSubt-long-meteor}
 & \UseMacro{ClsSubt-long-rouge-l}
\\
\UseMacro{GenModel}
 & \UseMacro{GenModel-short-bleu}
 & \UseMacro{GenModel-short-meteor}
 & \UseMacro{GenModel-short-rouge-l}
 & \UseMacro{GenModel-long-bleu}
 & \UseMacro{GenModel-long-meteor}
 & \UseMacro{GenModel-long-rouge-l}
\\
\UseMacro{DeepCom}
 & \UseMacro{DeepCom-short-bleu}
 & \UseMacro{DeepCom-short-meteor}
 & \UseMacro{DeepCom-short-rouge-l}
 & \UseMacro{DeepCom-long-bleu}
 & \UseMacro{DeepCom-long-meteor}
 & \UseMacro{DeepCom-long-rouge-l}
\\
\UseMacro{EditModel}
 & \UseMacro{EditModel-feat-rerank-short-bleu}
 & \UseMacro{EditModel-feat-rerank-short-meteor}
 & \UseMacro{EditModel-feat-rerank-short-rouge-l}
 & \UseMacro{EditModel-feat-rerank-long-bleu}
 & \UseMacro{EditModel-feat-rerank-long-meteor}
 & \UseMacro{EditModel-feat-rerank-long-rouge-l}
\\
\UseMacro{CopyGen-neg-remove-overlap}
 & \textbf{\UseMacro{CopyGen-neg-remove-overlap-short-bleu}}
 & \textbf{\UseMacro{CopyGen-neg-remove-overlap-short-meteor}}
 & \textbf{\UseMacro{CopyGen-neg-remove-overlap-short-rouge-l}}
 & \textbf{\UseMacro{CopyGen-neg-remove-overlap-long-bleu}}
 & \textbf{\UseMacro{CopyGen-neg-remove-overlap-long-meteor}}
 & \textbf{\UseMacro{CopyGen-neg-remove-overlap-long-rouge-l}}
\\
\bottomrule
\end{tabular}
\caption{\TableCaptionModelsResults}
\end{table*}

\begin{table*}[ht]
\small
\centering
\begin{tabular}{c|cllll}
\toprule
 \textbf{Dataset} & & \textbf{Model} & \textbf{BLEU-4} & \textbf{METEOR} &\textbf{ROUGE-L} \\
\midrule
\multirow{5}{*}{\Short{}} &
(1) &
{\small\SuperAwareGenWithSpecRerank{}}
  & \textbf{\UseMacro{CopyGen-neg-remove-overlap-short-bleu}}
  & \textbf{\UseMacro{CopyGen-neg-remove-overlap-short-meteor}}
  & \textbf{\UseMacro{CopyGen-neg-remove-overlap-short-rouge-l}}
\\
&
(2) &
{\hspace{0.2cm}\small\SuperAwareGenWOutUll{}}
 & \UseMacro{CopyGen-wo-ull-short-bleu}
 & \UseMacro{CopyGen-wo-ull-short-meteor}
 & \UseMacro{CopyGen-wo-ull-short-rouge-l}
\\
&
(3) &
{\hspace{0.2cm}\small\SuperAwareGenWOutSpecWOutRerank{}}
 & \UseMacro{CopyGen-short-bleu}
 & \UseMacro{CopyGen-short-meteor}$^{\eta}$
 & \UseMacro{CopyGen-short-rouge-l}$^{\zeta}$
\\
&
(4) &
{\hspace{0.2cm}\small\SuperAwareGenWOutFeatWOutRerankWOutSpec{}}
 & \UseMacro{CopyGen-feat-short-bleu}$^{\alpha}$
 & \UseMacro{CopyGen-feat-short-meteor}
 & \UseMacro{CopyGen-feat-short-rouge-l}$^{\zeta}$
\\
&
(5) &
{\hspace{0.2cm}\small\SuperAwareGenWOutClassName{}}
 & \UseMacro{CopyGen-class-short-bleu}$^{\alpha}$
 & \UseMacro{CopyGen-class-short-meteor}$^{\eta}$
 & \UseMacro{CopyGen-class-short-rouge-l}
\\
&
(6) &
{\hspace{0.2cm}\small\SuperAwareGenWOutComment{}}
 & \UseMacro{CopyGen-com-short-bleu}
 & \UseMacro{CopyGen-com-short-meteor}
 & \UseMacro{CopyGen-com-short-rouge-l}
\\

\midrule
\multirow{5}{*}{\Long{}} &
(1) &
{\small\SuperAwareGenWithSpecRerank{}}
  & \textbf{\UseMacro{CopyGen-neg-remove-overlap-long-bleu}}
  & \textbf{\UseMacro{CopyGen-neg-remove-overlap-long-meteor}}
  & \textbf{\UseMacro{CopyGen-neg-remove-overlap-long-rouge-l}}$^{\psi}$
\\
&
(2) &
{\hspace{0.2cm}\small\SuperAwareGenWOutUll{}}
 & \UseMacro{CopyGen-wo-ull-long-bleu}
 & \UseMacro{CopyGen-wo-ull-long-meteor}
 & \UseMacro{CopyGen-wo-ull-long-rouge-l}$^{\psi}$
\\
&
(3) &
{\hspace{0.2cm}\small\SuperAwareGenWOutSpecWOutRerank{}}
 & \UseMacro{CopyGen-long-bleu}$^{\delta\theta}$
 & \UseMacro{CopyGen-long-meteor}$^{\epsilon}$
 & \UseMacro{CopyGen-long-rouge-l}$^{\gamma\kappa}$
\\
&
(4) &
{\hspace{0.2cm}\small\SuperAwareGenWOutFeatWOutRerankWOutSpec{}}
 & \UseMacro{CopyGen-feat-long-bleu}$^{\theta}$
 & \UseMacro{CopyGen-feat-long-meteor}$^{\epsilon}$
 & \UseMacro{CopyGen-feat-long-rouge-l}$^{\gamma}$
\\
&
(5) &
{\hspace{0.2cm}\small\SuperAwareGenWOutClassName{}}
 & \UseMacro{CopyGen-class-long-bleu}$^{\delta}$
 & \UseMacro{CopyGen-class-long-meteor}
 & \UseMacro{CopyGen-class-long-rouge-l}$^{\kappa}$
\\
&
(6) &
{\hspace{0.2cm}\small\SuperAwareGenWOutComment{}}
 & \UseMacro{CopyGen-com-long-bleu}
 & \UseMacro{CopyGen-com-long-meteor}
 & \UseMacro{CopyGen-com-long-rouge-l}
\\

\bottomrule
\end{tabular}
\caption{\TableCaptionAblation}
\end{table*}

\renewcommand{\arraystretch}{0.9} %
\begin{table*}[ht]
\centering
\begin{tabular}{l}
\toprule
{
\begin{lstlisting}[language=java-pretty, numbers=none]
public class Object {
    /** Returns encoded form of the object */
    public byte[] getEncoded() {
        return encoding;}}

public class InfoAccessSyntax extends Object {
    /** ? */
    @Override public byte[] getEncoded() {   
        if (encoding == null) {
            encoding = ASN1.encode(this);}
        return encoding;}}

\end{lstlisting}}
\\

\small \textbf{Gold}: returns asn . 1 encoded form of this info access syntax .\\
\small \textbf{\Copy}:  returns encoded form of the object .\\
\small \textbf{\ClassNameSubst}: returns encoded form of the
info access syntax .\\ 
\small \textbf{\SuperAgnosticGen}: gets the encoded value .\\
\small \textbf{\DeepCom}: returns the byte 's value of the
encoding .\\
\small \textbf{\EditModel}: returns encoding of the object .\\ 
\small \textbf{\SuperAwareGen}: returns asn 1 encoded of the info
access syntax .\\
\bottomrule
\end{tabular}
\caption{\TableCaptionQualitativeExample}
\end{table*}

\section{Results}
\label{sec:eval}

In this section, we discuss performances of the various models. We
first present a quantitative analysis using automatic metrics
(Section~\ref{sec:eval:quantitative}), and then present a qualitative
analysis by inspecting the models' outputs
(Section~\ref{sec:eval:qualitative}).

\subsection{Quantitative Analysis}
\label{sec:eval:quantitative}

\MyPara{Metrics}
Following prior work in comment generation and code
summarization~\cite{iyer-etal-2016-summarizing, HuETAL19Deep, LiangAndZhu18Automatic, LeClairETAL19Neural}, we report metrics used to evaluate language generation tasks: BLEU-4~\cite{papineni-etal-2002-bleu}\footnote{Similar to prior work in code-to-language tasks~\cite{iyer-etal-2016-summarizing}, we report average sentence-level BLEU-4.}, METEOR~\cite{banerjee-lavie-2005-meteor} and
ROUGE-L~\cite{lin-2004-rouge}.
We report results as an average across three
random restarts (for learned models).

\MyPara{Results} In Table~\ref{tab:results}, we present results for
baselines and our \SuperAwareGen model.  We conducted statistical
significance testing through bootstrap
tests~\cite{Berg-KirkpatrickETAL12Empirical} under confidence level
95\%.

We first note that the \Copy baseline underperforms the majority of
other models (with the exception of \SuperAgnosticGen) across metrics
for both datasets, demonstrating that generating the comment for an
overriding method extends beyond simply repeating the overridden
method's comment.

Both \SuperAgnosticGen and \DeepCom, which do not have access to the class hierarchy, underperform the other three
approaches (besides \Copy) that do have access to this information,
including the rule-based \ClassNameSubst baseline. This underlines the
importance of leveraging the class hierarchy for generating
overriding method comments. \EditModel and \SuperAwareGen, the two
models which learn to exploit the class hierarchical context,
achieve even higher performance than \ClassNameSubst. We find
\SuperAwareGen to yield better performance than \EditModel which
suggests that generating a new comment by leveraging contextual information, specificity, and conformity extracted from the class hierarchy can lead to improved performance over simply learning to adapt \ParentComment{} to fit \ChildMethod{}.

Recall that \SuperAgnosticGen's core neural composition closely
matches that of \SuperAwareGen but does not have any information
pertaining to the class hierarchy. Hence, utilizing class hierarchy
can achieve more than 80\% improvement for each metric.

We observe that while the results are analogous, performance across
all metrics are consistently higher for \emph{first sentences} than
\emph{full descriptions}, indicating, not surprisingly, that the latter is a more
challenging task.

\MyPara{Ablation study}
\label{sec:ablation}
To evaluate the contribution of each of the components of
\SuperAwareGen, we perform an ablation study, with results shown in
Table~\ref{tab:ablt-study}. For both the first sentence and full
description tasks, we find that ablating each component deteriorates
performance across metrics; this effect is largely more evident on the
more challenging \xlong task. By comparing (1) and (2), we
demonstrate the importance of unlikelihood training. It gives
statistically significant improvements for all the metrics on the
\xshort task. The value of specificity
factoring is evident by comparing (2) and (3). Looking at (3) and (4),
we observe that ablating features drops performance w.r.t. BLEU-4 and
METEOR for the first sentence. Next, comparing (3) and (5), we see
that discarding the class name encoder deteriorates BLEU-4 and ROUGE-L
for the \xshort task and METEOR on the \xlong task. Finally, we
observe a sharp drop in performance when removing the parent comment
encoder (between (3) and (6)), underlining the importance of the
context provided by the \superclass comment for our proposed task.

\subsection{Qualitative Analysis}
\label{sec:eval:qualitative}

In Table~\ref{tab:qualitative-example}, we show the models' generated
\ChildComment{} (\xshort) for one overriding method
(\CodeIn{getEncoded()} in \CodeIn{InfoAccessSyntax}).  Copying
\ParentComment{} results in a generic and inappropriate
\ChildComment{}. \ClassNameSubst produces the correct
\ChildClassName{}, but the rest of the generated comment is not
specific enough. Without \ParentComment{} as context,
\SuperAgnosticGen and \DeepCom's generated comments are of low quality
and substantially diverged from \ParentComment{}. \EditModel performs
a rephrasing for \ParentComment{}, because its model architecture is
not designed towards improving the specificity of the
\ParentComment{}.  Our \SuperAwareGen generated a comment that is very
close to the gold \ChildComment{} with adequate specificity and
\compatibility.  We present more examples in Appendix~\ref{appendix:qualitative}.

\section{Related Work} 
\label{sec:related}

Most recent comment generation and code summarization approaches
entail encoding source code in various forms including the sequence of
tokens ~\cite{iyer-etal-2016-summarizing,
  AllamanisETAL16Convolutional, AhmadETAL20Transformer-based}, the
AST~\cite{LiangAndZhu18Automatic, AlonETAL19code2seq, HuETAL19Deep,
  CaiETAL20TAG}, or the data/control flow
graph~\cite{FernandesETAL19Structured, Liu2020AutomaticCS}, and then
decoding natural language output as a sequence of tokens.  However,
many of these approaches rely on the limited context provided by the
body of code alone.  In contrast, we additionally use context that is
external to the method, as we find this context to be critical for our
task of generating comments for \overriding methods.

\citet{LiuCallDependency} address the task of comment generation and
consider incorporating external context from the method call
dependency graph by encoding the method names as a sequence of
tokens. \citet{HaqueETAL20Improved} generate comments by encoding the
method signatures of all other methods in the same file. \citet{YuEtal20Towards} build a class-level contextual graph to relate all methods within the same class to a target method for which a comment is to be generated.
However, none of these approaches consider comments accompanying the methods extracted from the broader context. In this work, rather than the
context of the call graph or other methods within the same class or file, we
use the context provided by the class hierarchy, and we also extract a
comment (in the \superclass) from this context, which we found to be
one of the most critical components of our approach
(Section~\ref{sec:ablation}).

\citet{ZhaiETAL20CPC} propose a rule-based
approach for generating new comments by propagating comments from
various code elements such as methods and classes. Such a system can
generate a comment for the \subclass method by simply propagating the
\superclass comment (resembling our \Copy baseline) or doing simple
string replacements of class names (resembling our \ClassNameSubst
baseline). We instead \textit{learn} how to leverage \ParentComment{}
in our \SuperAwareGen model, which outperforms these baselines. \citet{AllamanisETAL15Suggesting} propose a log-bilinear neural language model for generating method names, which incorporates context from sibling methods and the superclass.
\citet{iyer-etal-2018-mapping} study the
reverse problem of generating a method given a natural language query
and context of the containing class. This includes class member
variables and the return types, names, and parameters of the other
methods in the class. We also use the class context (i.e.,
\ChildClassName{}), and context from another class,
specifically, through the \superclass.

\section{Conclusion}
\label{sec:conclusion}

This paper has introduced a new approach to automatic comment
generation for software by exploiting the hierarchical class structure
common in object-oriented programming languages. We propose a new framework which utilizes this hierarchy for contextual information as well as for learning notions of specificity and \compatibility, and we 
show that it outperforms existing state-of-the-art models on the novel
task of automatically commenting \overriding methods. We also
developed a new dataset
that allows training and testing such hierarchy-aware comment
generation models. Integrating this approach with the growing body of
work in machine learning and natural language processing for software
development will lead to the emergence of more intelligent,
effective software engineering environments.

\section*{Acknowledgments}
We thank Darko Marinov, Thomas Wei, and the anonymous reviewers for
their feedback on this work.  This research was partially supported by
the University of Texas at Austin Continuing Fellowship, the Bloomberg
Data Science Fellowship, a Google Faculty Research Award, and the US
National Science Foundation under Grant No. IIS-1850153.

\section*{Ethical Considerations}

Our dataset has been collected in a manner that is consistent with the
licenses provided from the sources (i.e., GitHub repositories).

Our technique is expected to work for software developers.  Our
technique will suggest a comment when developer override a method,
which the developer can review, modify, and add to the codebase.  In
case the technique makes a bad suggestion, the developer is expected
to detect it and refrain from using that suggestion.  We expect our
technique to improve developers' productivity and encourage developers
to write better documentations for code.  The technique should not be
used to automatically add comments to codebase without developers
first reviewing them.

We conducted experiments involving computation time/power, but we have
carefully chosen the number of times to repeat the experiments to both
ensure reproducibility of our research and avoid consuming excessive
energy.  We provide details of our computing platform and running time
in our appendix.

\bibliographystyle{acl_natbib}
\bibliography{bib}

\clearpage
\newpage
\appendix

\section{Incorporating AST Representations}
\label{appendix:AST}

\begin{table*}[ht]
\small
\centering
\begin{tabular}{c|rrr|rrr}
\toprule
\multirow{2}{*}{\textbf{Model}}
 &
\multicolumn{3}{c|}{\Short{}}
& 
\multicolumn{3}{c}{\Long{}}
\\
\cline{2-7}
& \textbf{BLEU-4} \Tstrut & \textbf{METEOR} &\textbf{ROUGE-L}
 & \textbf{BLEU-4} & \textbf{METEOR} &\textbf{ROUGE-L}\\
 \midrule
\SuperAwareGen{}
 & \textbf{\UseMacro{CopyGen-spec-rerank-short-bleu}}$^{\alpha}$
 & \textbf{\UseMacro{CopyGen-spec-rerank-short-meteor}}$^{\beta}$
 & \UseMacro{CopyGen-spec-rerank-short-rouge-l}$^{\zeta}$
 & \UseMacro{CopyGen-spec-rerank-long-bleu}$^{\eta}$
 & \textbf{\UseMacro{CopyGen-spec-rerank-long-meteor}}$^{\theta}$
 & \UseMacro{CopyGen-spec-rerank-long-rouge-l}$^{\gamma}$
\\
\SuperAwareGenWithAST{}
 & \UseMacro{AST-CopyGenCompatClfRerank-short-bleu}$^{\alpha}$
 & \UseMacro{AST-CopyGenCompatClfRerank-short-meteor}$^{\beta}$
 & \textbf{\UseMacro{AST-CopyGenCompatClfRerank-short-rouge-l}}$^{\zeta}$
 & \textbf{\UseMacro{AST-CopyGenCompatClfRerank-long-bleu}}$^{\eta}$
 & \UseMacro{AST-CopyGenCompatClfRerank-long-meteor}$^{\theta}$
 & \textbf{\UseMacro{AST-CopyGenCompatClfRerank-long-rouge-l}}$^{\gamma}$
\\
\bottomrule
\end{tabular}
\caption{Comparison of incorporating vs. not incorporating AST
  sequence into our model. The differences between the models with the
  same Greek letter suffix (and only those pairs) are \textbf{not}
  statistically significant.\label{tab:ast}}
\end{table*}

From Table~\ref{tab:results}, we observe that \DeepCom outperforms
\SuperAgnosticGen. Both models only have access to \ChildMethod{};
however, \DeepCom learns a representation for the AST corresponding to
\ChildMethod{} in addition to its sequence of code tokens. While our
model (\SuperAwareGen) equipped with context from the class hierarchy
without AST-level information outperforms \DeepCom, we are interested
in seeing whether using AST can further boost the performance of our
models.  For this, we introduce a fourth encoder into the
\SuperAwareGen model that learns a representation of \ChildMethod{}'s
AST.  Namely, we use \DeepCom's SBT algorithm to traverse
\ChildMethod{}'s AST and form an AST sequence. We then use a two-layer
bidirectional GRU encoder to encode this sequence. We present results
in Table~\ref{tab:ast}.  We find that incorporating the AST does not
yield statistically significant differences in performance.

\begin{table*}
\small
\centering
\begin{tabular}{c|rrr}
\toprule
\textbf{Model} & \textbf{Training Time[s]} & \textbf{Testing Time[s]} & \textbf{\#Parameters} \\
\midrule
\UseMacro{Copy} & \TableHeadNA & 6 & \TableHeadNA \\
\UseMacro{ClsSubt} & \TableHeadNA & 6 & \TableHeadNA \\
\UseMacro{GenModel} & 600 & 7,269 & 1.33M \\
\UseMacro{DeepCom} & 14,162 & 15,812 & 17.42M \\
\UseMacro{EditModel} & 1,147 & 9,384 & 2.19M \\
\UseMacro{CopyGen-neg-remove-overlap} & 919 & 8,143 & 3.33M\\
\bottomrule
\end{tabular}
\caption{Average training time, average testing time, and number of parameters of our \SuperAwareGen model and baseline models.\label{tab:runtime-num-params}}
\end{table*}

\section{Implementation and Experiments Details}
\label{appendix:reproducibility}

In this section, we provide all implementation and experiments details
for reproducibility that have not been covered in the main text.

\subsection{Hyper-Parameters}

The hyper-parameters of the \SuperAwareGen model
(Section~\ref{sec:CopyGen}) and the \SuperAgnosticGen model
(Section~\ref{sec:baselines}) are described below.  We use embedding
size $d_{comment}=d_{code}=64$.  Each encoder is a 2-layer
bidirectional GRU with hidden dimension $64$, and each decoder is a
2-layer unidirectional GRU with hidden dimension $128$.  Dropout rate
is set to $0.7$.  We optimize the negative log likelihood with Adam
optimizer using an learning rate of $0.001$.  In addition, we early
stop the training if the validation loss is not improving for
consequent $10$ epochs.  During inference, beam size of 20 is used.
For \DeepCom~\cite{HuETAL19Deep} and
\EditModel~\cite{PanthaplackelETAL20Learning}, we use the
hyper-parameters provided in the original papers.

We tune all of the hyper-parameters by maximizing the BLEU-4 on the
validation set.  The range of learning rates explored is \{0.1, 0.01,
0.001\}.  The range of beam sizes explored are the integers $\in [1,
  20]$.  The range of dropout rates explored is \{0.4, 0.6, 0.7,
0.8\}.  We also tried two optimization algorithms: SGD and Adam.  We
apply the same tuning process to the hyper-parameters for the other
models.

\subsection{Computing Environment}

We trained and tested all models on a super computer running Linux
operating system, where each model use one NVIDIA GTX 1080-TI GPU and
each four models share two Intel(R) Xeon(R) CPU E5-2620 CPUs and 128GB
of RAM.  We used Pytorch 1.2.0 to implement our \SuperAwareGen model
and the \SuperAgnosticGen baseline.  We trained and evaluated each
model in each evaluation setting \NumTrial times (each time with a
different random initialization of weights) and report average
metrics.

\subsection{Runtime and Number of Parameters}

Table~\ref{tab:runtime-num-params} shows our \SuperAwareGen model and baselines'
average runtime and number of parameters.

\section{Additional Qualitative Analysis Examples}
\label{appendix:qualitative}

In Table~\ref{tab:qualitative-example-app}, we show predictions for
various examples in the test set.

\renewcommand{\arraystretch}{0.8} %
\begin{table*}[ht]
\centering
\begin{tabular}{l}
\toprule
{
\begin{lstlisting}[language=java-pretty, numbers=none]
public class Cache {
    /** Empties the cache. */
    public synchronized void clear() {}
}

public class DiskBasedCache {
    /** ? */
    @Override public synchronized void clear() {
        File[] files = mRootDirectory.listFiles();
        if (files != null) {
            for (File file : files) {
                file.delete();
            }
        }
        mEntries.clear();
        mTotalSize = 0;
        VolleyLog.d("Cache cleared.");
    }
}

\end{lstlisting}}
\\
\\
\small \textbf{Gold}: clears the cache . deletes all cached files from disk .\\
\small \textbf{\Copy}: empties the cache .\\
\small \textbf{\ClassNameSubst}: empties the disk based cache .\\
\small \textbf{\SuperAgnosticGen}: removes all mappings from disk .\\
\small \textbf{\DeepCom}: deletes all cached files from disk .\\
\small \textbf{\EditModel}: clears the cache .\\ 
\small \textbf{\SuperAwareGen}: empties the cache . deletes all cached
files from disk .\\
\hline
{
\begin{lstlisting}[language=java-pretty, numbers=none]
public class Reader {
    /** Reset the stream. */
    public void reset() throws java.io.IOException {
        return;
    }
}

public class CharArrayReader {
    /** ? */
    @Override public void reset() throws IOException {
        synchronized (lock) {
            if (isClosed()) {
                throw new IOException();
            }
            pos = markedPos != -1 ? markedPos : 0;
        }
    }
}
\end{lstlisting}}
\\
\\
\small \textbf{Gold}: resets this reader ' s position to the last mark ( ) location .\\
\small \textbf{\Copy}: reset the stream .\\
\small \textbf{\ClassNameSubst}: empties the disk based cache .\\
\small \textbf{\SuperAgnosticGen}: removes all of the elements from this deque .\\
\small \textbf{\DeepCom}: overrides the reader to the last marked location ( ) and reset that it can be delegated to \# sent out ( ) .\\
\small \textbf{\EditModel}: clears the stream .\\ 
\small \textbf{\SuperAwareGen}: reset the stream to the tail of this reader .\\
\bottomrule
\end{tabular}
\caption{\TableCaptionQualitativeExampleApp}
\end{table*}

\end{document}

%% file: tables/numbers-dataset-metrics.tex
\DefMacro{raw-train-pairs}{26,206}
\DefMacro{raw-train-projects}{625}
\DefMacro{train-projects}{414}
\DefMacro{raw-valid-pairs}{1,361}
\DefMacro{raw-valid-projects}{32}
\DefMacro{valid-projects}{16}
\DefMacro{raw-test-pairs}{1,534}
\DefMacro{raw-test-projects}{58}
\DefMacro{test-projects}{41}
\DefMacro{long-train-numbers}{10,980}
\DefMacro{long-train-proj-num}{414}
\DefMacro{long-valid-numbers}{708}
\DefMacro{long-valid-proj-num}{16}
\DefMacro{long-test-numbers}{519}
\DefMacro{long-test-proj-num}{41}
\DefMacro{long-sup-comments-tokens}{37.688}
\DefMacro{long-sub-comments-tokens}{32.882}
\DefMacro{long-overall-comments-tokens}{35.285}
\DefMacro{long-sup-code-tokens}{38.74}
\DefMacro{long-sub-code-tokens}{106.736}
\DefMacro{longab-sup-comments-tokens}{34.361}
\DefMacro{longab-sub-comments-tokens}{32.155}
\DefMacro{longnonab-sup-comments-tokens}{43.575}
\DefMacro{longnonab-sub-comments-tokens}{34.168}
\DefMacro{num-abstract}{6,835}
\DefMacro{num-nonabstract}{3,719}
\DefMacro{short-train-numbers}{9,389}
\DefMacro{short-train-proj-num}{398}
\DefMacro{short-valid-numbers}{702}
\DefMacro{short-valid-proj-num}{16}
\DefMacro{short-test-numbers}{463}
\DefMacro{short-test-proj-num}{40}
\DefMacro{short-sup-comments-tokens}{15.969}
\DefMacro{short-sub-comments-tokens}{16.685}
\DefMacro{short-overall-comments-tokens}{16.327}
\DefMacro{short-sup-code-tokens}{38.481}
\DefMacro{short-sub-code-tokens}{106.275}
\DefMacro{shortab-sup-comments-tokens}{14.872}
\DefMacro{shortab-sub-comments-tokens}{16.386}
\DefMacro{shortnonab-sup-comments-tokens}{17.987}
\DefMacro{shortnonab-sub-comments-tokens}{17.236}

\DefMacro{old-long-train-numbers}{12,118}
\DefMacro{old-long-valid-numbers}{639}
\DefMacro{old-long-test-numbers}{771}

\DefMacro{old-short-test-numbers}{614}
\DefMacro{old-short-train-numbers}{10,804}
\DefMacro{old-short-valid-numbers}{598}

%% file: tables/numbers-comlen.tex
\DefMacro{hie-sup}{16.4}
\DefMacro{hie-sub}{13.8}
\DefMacro{hie-overall}{15.1}

%% file: tables/numbers-copy-results.tex
\DefMacro{Copy-short-bleu}{24.228}
\DefMacro{Copy-short-meteor}{19.845}
\DefMacro{Copy-short-rouge-l}{43.352}
\DefMacro{Copy-long-bleu}{20.623}
\DefMacro{Copy-long-meteor}{18.757}
\DefMacro{Copy-long-rouge-l}{40.846}

%% file: tables/numbers-deepcom-results.tex
\DefMacro{DeepCom-short-bleu}{26.073}
\DefMacro{DeepCom-short-meteor}{22.256}
\DefMacro{DeepCom-short-rouge-l}{45.037}
\DefMacro{DeepCom-long-bleu}{21.812}
\DefMacro{DeepCom-long-meteor}{20.779}
\DefMacro{DeepCom-long-rouge-l}{40.257}

%% file: tables/numbers-clssubt-results.tex
\DefMacro{ClsSubt-short-bleu}{26.764}
\DefMacro{ClsSubt-short-meteor}{22.496}
\DefMacro{ClsSubt-short-rouge-l}{45.632}
\DefMacro{ClsSubt-long-bleu}{24.215}
\DefMacro{ClsSubt-long-meteor}{21.442}
\DefMacro{ClsSubt-long-rouge-l}{42.014}

%% file: tables/numbers-genmodel-results.tex
\DefMacro{GenModel-long-bleu}{9.003}
\DefMacro{GenModel-long-meteor}{9.136}
\DefMacro{GenModel-long-rouge-l}{24.379}
\DefMacro{GenModel-short-bleu}{15.389}
\DefMacro{GenModel-short-meteor}{13.122}
\DefMacro{GenModel-short-rouge-l}{29.453}

%% file: tables/numbers-copygen-wo-ull-results.tex
\DefMacro{CopyGen-wo-ull-long-bleu}{27.307}
\DefMacro{CopyGen-wo-ull-long-meteor}{26.323}
\DefMacro{CopyGen-wo-ull-long-rouge-l}{48.272}
\DefMacro{CopyGen-wo-ull-short-bleu}{35.754}
\DefMacro{CopyGen-wo-ull-short-meteor}{31.272}
\DefMacro{CopyGen-wo-ull-short-rouge-l}{53.732}

%% file: tables/numbers-editmodel-results.tex
\DefMacro{EditModel-long-bleu}{22.232}
\DefMacro{EditModel-long-meteor}{21.729}
\DefMacro{EditModel-long-rouge-l}{45.186}
\DefMacro{EditModel-short-bleu}{31.666}
\DefMacro{EditModel-short-meteor}{27.761}
\DefMacro{EditModel-short-rouge-l}{50.010}

%% file: tables/numbers-copygen-results.tex
\DefMacro{CopyGen-long-bleu}{23.681}
\DefMacro{CopyGen-long-meteor}{23.247}
\DefMacro{CopyGen-long-rouge-l}{46.435}
\DefMacro{CopyGen-short-bleu}{34.610}
\DefMacro{CopyGen-short-meteor}{29.962}
\DefMacro{CopyGen-short-rouge-l}{52.588}

%% file: tables/numbers-ensemble-results.tex
\DefMacro{Ensemble-short-acc}{2.544}
\DefMacro{Ensemble-short-bleu}{36.369}
\DefMacro{Ensemble-short-meteor}{31.928}
\DefMacro{Ensemble-short-sari}{39.183}
\DefMacro{Ensemble-short-gleu}{28.441}
\DefMacro{Ensemble-short-rouge-l}{53.844}
\DefMacro{Ensemble-long-acc}{1.789}
\DefMacro{Ensemble-long-bleu}{26.689}
\DefMacro{Ensemble-long-meteor}{25.769}
\DefMacro{Ensemble-long-sari}{32.085}
\DefMacro{Ensemble-long-gleu}{20.031}
\DefMacro{Ensemble-long-rouge-l}{47.481}

\DefMacro{meta-classifier-short-acc}{42.5}
\DefMacro{meta-classifier-long-acc}{38.4}

%% file: tables/numbers-ablation-results.tex
\DefMacro{CopyGen-feat-short-acc}{0.000}
\DefMacro{CopyGen-feat-short-bleu}{33.715}
\DefMacro{CopyGen-feat-short-meteor}{29.029}
\DefMacro{CopyGen-feat-short-sari}{32.929}
\DefMacro{CopyGen-feat-short-gleu}{20.736}
\DefMacro{CopyGen-feat-short-rouge-l}{52.688}
\DefMacro{CopyGen-feat-long-acc}{0.000}
\DefMacro{CopyGen-feat-long-bleu}{24.303}
\DefMacro{CopyGen-feat-long-meteor}{23.486}
\DefMacro{CopyGen-feat-long-sari}{32.921}
\DefMacro{CopyGen-feat-long-gleu}{14.424}
\DefMacro{CopyGen-feat-long-rouge-l}{46.722}
\DefMacro{CopyGen-class-short-acc}{0.000}
\DefMacro{CopyGen-class-short-bleu}{33.969}
\DefMacro{CopyGen-class-short-meteor}{28.984}
\DefMacro{CopyGen-class-short-sari}{33.114}
\DefMacro{CopyGen-class-short-gleu}{19.971}
\DefMacro{CopyGen-class-short-rouge-l}{52.098}
\DefMacro{CopyGen-class-long-acc}{0.000}
\DefMacro{CopyGen-class-long-bleu}{23.745}
\DefMacro{CopyGen-class-long-meteor}{22.887}
\DefMacro{CopyGen-class-long-sari}{33.235}
\DefMacro{CopyGen-class-long-gleu}{15.309}
\DefMacro{CopyGen-class-long-rouge-l}{46.335}
\DefMacro{CopyGen-com-short-acc}{0.000}
\DefMacro{CopyGen-com-short-bleu}{22.694}
\DefMacro{CopyGen-com-short-meteor}{21.209}
\DefMacro{CopyGen-com-short-sari}{33.764}
\DefMacro{CopyGen-com-short-gleu}{18.094}
\DefMacro{CopyGen-com-short-rouge-l}{41.948}
\DefMacro{CopyGen-com-long-acc}{0.000}
\DefMacro{CopyGen-com-long-bleu}{13.890}
\DefMacro{CopyGen-com-long-meteor}{15.115}
\DefMacro{CopyGen-com-long-sari}{33.964}
\DefMacro{CopyGen-com-long-gleu}{12.128}
\DefMacro{CopyGen-com-long-rouge-l}{33.907}

\DefMacro{AST-CopyGen-short-acc}{0.072}
\DefMacro{AST-CopyGen-short-bleu}{34.842}
\DefMacro{AST-CopyGen-short-meteor}{30.200}
\DefMacro{AST-CopyGen-short-sari}{33.272}
\DefMacro{AST-CopyGen-short-gleu}{23.674}
\DefMacro{AST-CopyGen-short-rouge-l}{52.768}
\DefMacro{AST-CopyGen-long-acc}{0.000}
\DefMacro{AST-CopyGen-long-bleu}{24.791}
\DefMacro{AST-CopyGen-long-meteor}{24.031}
\DefMacro{AST-CopyGen-long-sari}{33.087}
\DefMacro{AST-CopyGen-long-gleu}{15.480}
\DefMacro{AST-CopyGen-long-rouge-l}{47.146}

%% file: tables/numbers-copygen-spec-results.tex
\DefMacro{CopyGen-spec-long-bleu}{24.636}
\DefMacro{CopyGen-spec-long-meteor}{23.978}
\DefMacro{CopyGen-spec-long-rouge-l}{46.681}
\DefMacro{CopyGen-spec-short-bleu}{35.504}
\DefMacro{CopyGen-spec-short-meteor}{30.726}
\DefMacro{CopyGen-spec-short-rouge-l}{53.234}

%% file: tables/numbers-copygen-neg-remove-overlap-results.tex
\DefMacro{CopyGen-neg-remove-overlap-long-bleu}{27.461}
\DefMacro{CopyGen-neg-remove-overlap-long-meteor}{26.361}
\DefMacro{CopyGen-neg-remove-overlap-long-rouge-l}{48.698}
\DefMacro{CopyGen-neg-remove-overlap-short-bleu}{36.408}
\DefMacro{CopyGen-neg-remove-overlap-short-meteor}{31.955}
\DefMacro{CopyGen-neg-remove-overlap-short-rouge-l}{53.776}

%% file: tables/numbers-copygencompatclfrerank-results.tex
\DefMacro{CopyGen-spec-rerank-long-bleu}{27.719}
\DefMacro{CopyGen-spec-rerank-long-meteor}{26.908}
\DefMacro{CopyGen-spec-rerank-long-rouge-l}{49.352}
\DefMacro{CopyGen-spec-rerank-short-bleu}{35.899}
\DefMacro{CopyGen-spec-rerank-short-meteor}{31.131}
\DefMacro{CopyGen-spec-rerank-short-rouge-l}{53.548}

%% file: tables/numbers-specificity.tex
\DefMacro{sub-niwf-short-train-AVG}{0.249}
\DefMacro{sup-niwf-short-train-AVG}{0.174}
\DefMacro{sub-niwf-short-train-SUM}{2337.045}
\DefMacro{sup-niwf-short-train-SUM}{1637.676}
\DefMacro{sub-niwf-short-train-MAX}{1.000}
\DefMacro{sup-niwf-short-train-MAX}{1.000}
\DefMacro{sub-niwf-short-train-MIN}{0.000}
\DefMacro{sup-niwf-short-train-MIN}{0.000}
\DefMacro{sub-niwf-short-train-MEDIAN}{0.116}
\DefMacro{sup-niwf-short-train-MEDIAN}{0.078}
\DefMacro{sub-niwf-short-train-STDEV}{0.292}
\DefMacro{sup-niwf-short-train-STDEV}{0.226}
\DefMacro{niwf-significance-level-short-train}{0.0000}
\DefMacro{sub-niwf-short-valid-AVG}{1.119}
\DefMacro{sup-niwf-short-valid-AVG}{1.015}
\DefMacro{sub-niwf-short-valid-SUM}{785.728}
\DefMacro{sup-niwf-short-valid-SUM}{712.666}
\DefMacro{sub-niwf-short-valid-MAX}{2.002}
\DefMacro{sup-niwf-short-valid-MAX}{2.002}
\DefMacro{sub-niwf-short-valid-MIN}{0.000}
\DefMacro{sup-niwf-short-valid-MIN}{0.002}
\DefMacro{sub-niwf-short-valid-MEDIAN}{1.000}
\DefMacro{sup-niwf-short-valid-MEDIAN}{0.666}
\DefMacro{sub-niwf-short-valid-STDEV}{0.835}
\DefMacro{sup-niwf-short-valid-STDEV}{0.862}
\DefMacro{niwf-significance-level-short-valid}{0.0003}
\DefMacro{sub-niwf-short-test-AVG}{0.506}
\DefMacro{sup-niwf-short-test-AVG}{0.324}
\DefMacro{sub-niwf-short-test-SUM}{234.175}
\DefMacro{sup-niwf-short-test-SUM}{150.201}
\DefMacro{sub-niwf-short-test-MAX}{2.002}
\DefMacro{sup-niwf-short-test-MAX}{2.002}
\DefMacro{sub-niwf-short-test-MIN}{0.002}
\DefMacro{sup-niwf-short-test-MIN}{0.002}
\DefMacro{sub-niwf-short-test-MEDIAN}{0.075}
\DefMacro{sup-niwf-short-test-MEDIAN}{0.067}
\DefMacro{sub-niwf-short-test-STDEV}{0.757}
\DefMacro{sup-niwf-short-test-STDEV}{0.585}
\DefMacro{niwf-significance-level-short-test}{0.0000}
\DefMacro{sub-niwf-long-train-AVG}{0.244}
\DefMacro{sup-niwf-long-train-AVG}{0.181}
\DefMacro{sub-niwf-long-train-SUM}{2677.344}
\DefMacro{sup-niwf-long-train-SUM}{1992.461}
\DefMacro{sub-niwf-long-train-MAX}{1.000}
\DefMacro{sup-niwf-long-train-MAX}{1.000}
\DefMacro{sub-niwf-long-train-MIN}{0.000}
\DefMacro{sup-niwf-long-train-MIN}{0.000}
\DefMacro{sub-niwf-long-train-MEDIAN}{0.110}
\DefMacro{sup-niwf-long-train-MEDIAN}{0.082}
\DefMacro{sub-niwf-long-train-STDEV}{0.292}
\DefMacro{sup-niwf-long-train-STDEV}{0.233}
\DefMacro{niwf-significance-level-long-train}{0.0000}
\DefMacro{sub-niwf-long-valid-AVG}{0.919}
\DefMacro{sup-niwf-long-valid-AVG}{0.910}
\DefMacro{sub-niwf-long-valid-SUM}{650.382}
\DefMacro{sup-niwf-long-valid-SUM}{644.092}
\DefMacro{sub-niwf-long-valid-MAX}{2.001}
\DefMacro{sup-niwf-long-valid-MAX}{2.001}
\DefMacro{sub-niwf-long-valid-MIN}{0.001}
\DefMacro{sup-niwf-long-valid-MIN}{0.002}
\DefMacro{sub-niwf-long-valid-MEDIAN}{0.499}
\DefMacro{sup-niwf-long-valid-MEDIAN}{0.499}
\DefMacro{sub-niwf-long-valid-STDEV}{0.841}
\DefMacro{sup-niwf-long-valid-STDEV}{0.850}
\DefMacro{niwf-significance-level-long-valid}{0.6031}
\DefMacro{sub-niwf-long-test-AVG}{0.448}
\DefMacro{sup-niwf-long-test-AVG}{0.332}
\DefMacro{sub-niwf-long-test-SUM}{232.275}
\DefMacro{sup-niwf-long-test-SUM}{172.292}
\DefMacro{sub-niwf-long-test-MAX}{2.001}
\DefMacro{sup-niwf-long-test-MAX}{2.001}
\DefMacro{sub-niwf-long-test-MIN}{0.002}
\DefMacro{sup-niwf-long-test-MIN}{0.002}
\DefMacro{sub-niwf-long-test-MEDIAN}{0.073}
\DefMacro{sup-niwf-long-test-MEDIAN}{0.061}
\DefMacro{sub-niwf-long-test-STDEV}{0.704}
\DefMacro{sup-niwf-long-test-STDEV}{0.597}
\DefMacro{niwf-significance-level-long-test}{0.0008}

%% file: tables/numbers-compat-clf.tex
\DefMacro{compat-clf-short-valid-clf_acc}{71.3}
\DefMacro{compat-clf-short-valid-clf_f1}{0.732}
\DefMacro{compat-clf-short-valid-clf_precision}{68.6}
\DefMacro{compat-clf-short-valid-clf_recall}{78.5}
\DefMacro{compat-clf-short-test-clf_acc}{75.5}
\DefMacro{compat-clf-short-test-clf_f1}{0.764}
\DefMacro{compat-clf-short-test-clf_precision}{73.6}
\DefMacro{compat-clf-short-test-clf_recall}{79.5}
\DefMacro{compat-clf-long-valid-clf_acc}{62.9}
\DefMacro{compat-clf-long-valid-clf_f1}{0.672}
\DefMacro{compat-clf-long-valid-clf_precision}{60.2}
\DefMacro{compat-clf-long-valid-clf_recall}{76.1}
\DefMacro{compat-clf-long-test-clf_acc}{71.5}
\DefMacro{compat-clf-long-test-clf_f1}{0.742}
\DefMacro{compat-clf-long-test-clf_precision}{67.8}
\DefMacro{compat-clf-long-test-clf_recall}{81.9}

%% file: tables/numbers-ast-copygencompatclfrerank-results.tex
\DefMacro{AST-CopyGenCompatClfRerank-long-bleu}{27.908}
\DefMacro{AST-CopyGenCompatClfRerank-long-meteor}{26.360}
\DefMacro{AST-CopyGenCompatClfRerank-long-rouge-l}{49.935}
\DefMacro{AST-CopyGenCompatClfRerank-short-bleu}{35.837}
\DefMacro{AST-CopyGenCompatClfRerank-short-meteor}{30.909}
\DefMacro{AST-CopyGenCompatClfRerank-short-rouge-l}{53.749}

%% file: tables/numbers-editmodel-feat-rerank-results.tex
\DefMacro{EditModel-feat-rerank-long-bleu}{23.734}
\DefMacro{EditModel-feat-rerank-long-meteor}{22.571}
\DefMacro{EditModel-feat-rerank-long-rouge-l}{44.359}
\DefMacro{EditModel-feat-rerank-short-bleu}{31.755}
\DefMacro{EditModel-feat-rerank-short-meteor}{26.926}
\DefMacro{EditModel-feat-rerank-short-rouge-l}{46.939}

%% file: figures/model-architecture.tex
\definecolor{highlightcolor}{cmyk}{0,.66,.55,.47}  % red
\definecolor{featcolor}{cmyk}{.23,0,1,.29}  % applegreen
\definecolor{speccolor}{cmyk}{0.9,0,.88,.5}  % dartmouthgreen

\begin{tikzpicture}[
    every node/.style={font=\small},
  ]

  \tikzset{box/.style={
      text=white,
      draw=highlightcolor,
      fill=highlightcolor,
      font=\small,
  }}

  \tikzset{textbox/.style={
      rounded corners,
      draw=black,
      fill=blue!10,
      font=\small,
  }}

  \tikzset{textbox io/.style={
      draw=none,
      font=\small,
      text height=2mm,
      text depth=.4mm,
  }}

  \tikzset{nnode/.style={
      draw=black,
      minimum width=1mm,
      minimum height=1mm,
  }}

  \tikzset{anno/.style={
      font=\small,
  }}

  \node[coordinate] (c-anchor) at (0,0) {};
  \node[coordinate] (c-w-enc) at (-45mm,0) {};
  \node[coordinate] (c-w-dec) at (10mm,0) {};

  \node[coordinate] (c-w-enc1) [above = 10mm of c-w-enc] {};
  \node[coordinate] (c-w-enc2) at (c-w-enc) {};
  \node[coordinate] (c-w-enc3) [below = 10mm of c-w-enc] {};

  \newcommand{\wSepNNode}{8mm}

  % Encoder 1
  \node[anno] (n-enc) [below right = 8mm and 6mm of c-w-enc3] {ENCODERS};
  \node[textbox] (n-in1) [left = 3mm of c-w-enc1] {\ChildMethod{}};
  \node[nnode] (n-enc1-1) [right = 0 of c-w-enc1] {};
  \node[nnode] (n-enc1-2) [right = \wSepNNode of n-enc1-1] {};
  \node[textbox, draw=none, fill=none] (n-enc1-3) [right = \wSepNNode of n-enc1-2] {$\cdots$};
  \node[nnode] (n-enc1-4) [right = \wSepNNode of n-enc1-3] {};

  \draw[->] ([yshift=.3mm]n-enc1-1.east) -- ([yshift=.3mm]n-enc1-2.west);
  \draw[->] ([yshift=-.3mm]n-enc1-2.west) -- ([yshift=-.3mm]n-enc1-1.east);
  \draw[->] ([yshift=.3mm]n-enc1-2.east) -- ([yshift=.3mm]n-enc1-3.west);
  \draw[->] ([yshift=-.3mm]n-enc1-3.west) -- ([yshift=-.3mm]n-enc1-2.east);
  \draw[->] ([yshift=.3mm]n-enc1-3.east) -- ([yshift=.3mm]n-enc1-4.west);
  \draw[->] ([yshift=-.3mm]n-enc1-4.west) -- ([yshift=-.3mm]n-enc1-3.east);

  \node[textbox io] (n-in1-1) [below = 2mm of n-enc1-1] {public};
  \node[textbox io] (n-in1-2) [below = 2mm of n-enc1-2] {byte};
  \node[textbox io] (n-in1-4) [below = 2mm of n-enc1-4] {\}};

  \draw[->] (n-in1-1) -- (n-enc1-1);
  \draw[->] (n-in1-2) -- (n-enc1-2);
  \draw[->] (n-in1-4) -- (n-enc1-4);

  % Encoder 2
  \node[textbox] (n-in2) [left = 3mm of c-w-enc2] {\ChildClassName{}};
  \node[nnode] (n-enc2-1) [right = 0 of c-w-enc2] {};
  \node[nnode] (n-enc2-2) [right = \wSepNNode of n-enc2-1] {};
  \node[textbox, draw=none, fill=none] (n-enc2-3) [right = \wSepNNode of n-enc2-2] {$\cdots$};
  \node[nnode] (n-enc2-4) [right = \wSepNNode of n-enc2-3] {};

  \draw[->] ([yshift=.3mm]n-enc2-1.east) -- ([yshift=.3mm]n-enc2-2.west);
  \draw[->] ([yshift=-.3mm]n-enc2-2.west) -- ([yshift=-.3mm]n-enc2-1.east);
  \draw[->] ([yshift=.3mm]n-enc2-2.east) -- ([yshift=.3mm]n-enc2-3.west);
  \draw[->] ([yshift=-.3mm]n-enc2-3.west) -- ([yshift=-.3mm]n-enc2-2.east);
  \draw[->] ([yshift=.3mm]n-enc2-3.east) -- ([yshift=.3mm]n-enc2-4.west);
  \draw[->] ([yshift=-.3mm]n-enc2-4.west) -- ([yshift=-.3mm]n-enc2-3.east);

  \node[textbox io] (n-in2-1) [below = 2mm of n-enc2-1] {info};
  \node[textbox io] (n-in2-2) [below = 2mm of n-enc2-2] {access};
  \node[textbox io] (n-in2-4) [below = 2mm of n-enc2-4] {syntax};

  \draw[->] (n-in2-1) -- (n-enc2-1);
  \draw[->] (n-in2-2) -- (n-enc2-2);
  \draw[->] (n-in2-4) -- (n-enc2-4);

  % Encoder 3
  \node[textbox] (n-in3) [left = 3mm of c-w-enc3] {\ParentComment{}};
  \node[nnode] (n-enc3-1) [right = 0 of c-w-enc3] {};
  \node[nnode] (n-enc3-2) [right = \wSepNNode of n-enc3-1] {};
  \node[textbox, draw=none, fill=none] (n-enc3-3) [right = \wSepNNode of n-enc3-2] {$\cdots$};
  \node[nnode] (n-enc3-4) [right = \wSepNNode of n-enc3-3] {};

  \draw[->] ([yshift=.3mm]n-enc3-1.east) -- ([yshift=.3mm]n-enc3-2.west);
  \draw[->] ([yshift=-.3mm]n-enc3-2.west) -- ([yshift=-.3mm]n-enc3-1.east);
  \draw[->] ([yshift=.3mm]n-enc3-2.east) -- ([yshift=.3mm]n-enc3-3.west);
  \draw[->] ([yshift=-.3mm]n-enc3-3.west) -- ([yshift=-.3mm]n-enc3-2.east);
  \draw[->] ([yshift=.3mm]n-enc3-3.east) -- ([yshift=.3mm]n-enc3-4.west);
  \draw[->] ([yshift=-.3mm]n-enc3-4.west) -- ([yshift=-.3mm]n-enc3-3.east);

  \node[textbox io] (n-in3-1) [below = 2mm of n-enc3-1] {returns};
  \node[textbox io,xshift=.5mm] (n-in3-2) [below = 2mm of n-enc3-2] {encoded};
  \node[textbox io] (n-in3-4) [below = 2mm of n-enc3-4] {object};

  \draw[->] (n-in3-1) -- (n-enc3-1);
  \draw[->] ([xshift=-.5mm]n-in3-2.north) -- (n-enc3-2);
  \draw[->] (n-in3-4) -- (n-enc3-4);

  % FC layer
  \draw[draw=black] ([yshift=-4mm]c-anchor) -- ([yshift=4mm]c-anchor) -- ([xshift=7mm]c-anchor) -- cycle;
  \node[anno] [below right = 5mm and -3mm of c-anchor] {};
  \draw[->,>=stealth',rounded corners] (n-enc1-4) -- +(3mm,0) -- ([yshift=2mm]c-anchor);
  \draw[->,>=stealth',rounded corners] (n-enc2-4) -- +(3mm,0) -- ([yshift=0]c-anchor);
  \draw[->,>=stealth',rounded corners] (n-enc3-4) -- +(3mm,0) -- ([yshift=-2mm]c-anchor);

  % Decoder

  \node[nnode] (n-dec-1) [right = 0 of c-w-dec] {};
  \node[nnode] (n-dec-2) [right = \wSepNNode of n-dec-1] {};
  \node[nnode] (n-dec-3) [right = \wSepNNode of n-dec-2] {};
  \node[nnode] (n-dec-4) [right = \wSepNNode of n-dec-3] {};

  \node[textbox io] (n-out-0) [below = 2mm of n-dec-1] {$\langle$BOS$\rangle$};
  \node[textbox io] (n-out-1) [above = 2mm of n-dec-1] {returns};
  \node[textbox io] (n-out-2) [above = 2mm of n-dec-2] {astn};
  \node[textbox io] (n-out-3) [above = 2mm of n-dec-3] {$\dots$};
  \node[textbox io] (n-out-4) [above = 2mm of n-dec-4] {$\langle$EOS$\rangle$};

  \node[textbox] (n-out) [right = 2mm of n-dec-4] {\ChildComment{}};
  \node[] (n-dec) [below right = 10mm and 8mm of c-w-dec] {DECODER};

  \draw[->] ([xshift=7mm]c-anchor) -- (n-dec-1);
  \draw[->] (n-out-0) -- (n-dec-1);
  \draw[->] (n-dec-1) -- (n-out-1);
  \draw[->] (n-dec-2) -- (n-out-2);
  \draw[->] (n-dec-3) -- (n-out-3);
  \draw[->] (n-dec-4) -- (n-out-4);

  \draw[->] (n-dec-1.east) -- (n-dec-2.west);
  \draw[->] (n-dec-2.east) -- (n-dec-3.west);
  \draw[->] (n-dec-3.east) -- (n-dec-4.west);

  \draw[->] (n-out-1.east) -- ++(.5mm,0) -- ++(0,-9mm) -| (n-dec-2.south);
  \draw[->] (n-out-2.east) -- ++(1mm,0) -- ++(0,-9mm) -| (n-dec-3.south);
  \draw[->] (n-out-3.east) -- ++(.5mm,0) -- ++(0,-9mm) -| (n-dec-4.south);

  % Spec
  \node[textbox, rounded corners, draw=black, fill=highlightcolor!10] (n-spec-level) [below right = 22mm and -8mm of c-anchor] {
  \begin{tabular}{c|c}
    \makecell[c]{\textbf{specificity level}\\\textbf{embedding}} &
    \makecell[c]{
      training time: $\mathtt{specificity\_level}(\ChildComment)$\\
      inference time: $K$ (maximum specificity level)
    }
  \end{tabular}
  };

  \node[coordinate] (c-spec-level) at (c-anchor |- n-spec-level.north) [xshift=0] {};

  \draw[->] (c-spec-level) .. controls ++(0, 14mm) .. (n-dec-1.south);
  \draw[->] (c-spec-level) .. controls ++(0, 8mm) .. (n-dec-2.south);
  \draw[->] (c-spec-level) .. controls ++(0, 8mm) .. (n-dec-3.south);
  \draw[->] (c-spec-level) .. controls ++(0, 8mm) .. (n-dec-4.south);

\end{tikzpicture}

%% file: tables/table-dataset-metrics.tex
%% Automatically generated by pyutil.latex 

\begin{table}
\begin{small}
\begin{center}
\begin{tabular}{l|rrr}
\toprule
 & \textbf{\UseMacro{TableHead-train}}
 & \textbf{\UseMacro{TableHead-valid}}
 & \textbf{\UseMacro{TableHead-test}}
\\
\midrule
\TableHeadNumProjects
 & \UseMacro{train-projects}
 & \UseMacro{valid-projects}
 & \UseMacro{test-projects}
\\
\midrule
\TableHeadBeforePreprocessing
 & \UseMacro{raw-train-pairs}
 & \UseMacro{raw-valid-pairs}
 & \UseMacro{raw-test-pairs}
\\
\midrule
\UseMacro{TableHead-short}
 & \UseMacro{short-train-numbers}
 & \UseMacro{short-valid-numbers}
 & \UseMacro{short-test-numbers}
\\
\UseMacro{TableHead-long}
 & \UseMacro{long-train-numbers}
 & \UseMacro{long-valid-numbers}
 & \UseMacro{long-test-numbers}
\\
\bottomrule
\end{tabular}
\end{center}
\end{small}
\caption{\TableCaptionDatasetMetrics}
\end{table}

%% file: tables/table-dataset-metrics-aux.tex
%% Automatically generated by pyutil.latex 

\begin{table}
\begin{small}
\begin{center}
\begin{tabular}{l|rr}
\toprule
 & \TableHeadAvgLenParentComment
 & \TableHeadAvgLenChildComment
\\
\midrule
\UseMacro{TableHead-short}
 & \UseMacro{short-sup-comments-tokens}
 & \UseMacro{short-sub-comments-tokens}
\\
\UseMacro{TableHead-long}
 & \UseMacro{long-sup-comments-tokens}
 & \UseMacro{long-sub-comments-tokens}
\\
\bottomrule
\end{tabular}
\end{center}
\end{small}
\caption{\TableCaptionDatasetMetricsAux}
\end{table}